%% file: main_arxiv.tex
\documentclass{article}

\usepackage{natbib}
\usepackage[left=1.25in, top=1in, bottom=1in, right=1.25in]{geometry}
\usepackage[utf8]{inputenc} 
\usepackage[T1]{fontenc}    
\usepackage{url}            
\usepackage{booktabs}       
\usepackage{amsfonts}       
\usepackage{nicefrac}       
\usepackage{microtype}      
\usepackage{amsmath}
\usepackage{amsthm}
\usepackage{graphicx}

\usepackage{subfig}
\usepackage[inline]{enumitem}
\usepackage{stmaryrd}
\usepackage[mathscr]{eucal}
\usepackage{amsmath}
\usepackage[export]{adjustbox}

\usepackage{colortbl}

\usepackage[dvipsnames]{xcolor}

\newcommand\gui[1]{}
\newcommand\aditya[1]{}
\newcommand{\best}[1]{\cellcolor{gray!25}{#1}}

\newcommand{\leaves}{\operatorname{leaves}}
\newcommand{\lca}{\operatorname{lca}}
\newcommand{\Rev}{\operatorname{Rev}}
\newcommand{\Cost}{\operatorname{Cost}}
\newcommand{\MaxRev}{\operatorname{MaxRev}}

\newcommand{\wbar}{\overline{w}}
\newcommand{\Merge}{\operatorname{Merge}}
\newcommand{\Node}{\operatorname{Node}}
\newcommand{\OTD}{\operatorname{OTD}}
\newtheorem{fact}{Fact}
\newtheorem{assumption}{Assumption}
\newtheorem{definition}{Definition}
\newtheorem{example}{Example}

\newtheorem{theorem}{Theorem}

\usepackage{algorithm}
\usepackage{algorithmic}

\usepackage[framemethod=tikz]{mdframed}
\newmdenv[
  linecolor=cyan,
  linewidth=2pt,
  roundcorner=5pt,
  innertopmargin=0pt,
  innerbottommargin=0pt,
]{myframe}

\title{Online hierarchical clustering approximations}

\author{%
  Anand Rajagopalan \\ 
  Google Research\\
  \texttt{anandbr@google.com} \\
  \and
  Aditya Krishna Menon \\ 
  Google Research\\
  \texttt{adityakmenon@google.com} \\
  \and
  Qin Cao \\ 
  Google Research\\
  \texttt{qincao@google.com} \\
  \and
  Gui Citovsky \\ 
  Google Research\\
  \texttt{gcitovsky@google.com} \\
  \and
  Baris Sumengen \\ 
  Google Research\\
  \texttt{sumengen@google.com} \\
  \and
  Sanjiv Kumar \\ 
  Google Research\\
  \texttt{sanjivk@google.com} \\
}

\begin{document}

\maketitle

\begin{abstract}
\input{abstract}
\end{abstract}

\section{Introduction}
\label{Introdution}
\input{intro}

\section{Background and related work}
\label{Background}

\input{background}

\section{Online top down clustering (OTD)}
\label{OTD}
\input{otd}

\section{Online HAC (OHAC)}
\label{ohac}
\input{ohac}

\section{Experimental results}
\label{Evaluations}
\input{evaluation}

\vspace{-1mm}
\section{Conclusion}
\label{Conclusion}
\input{conclusion}

\clearpage
\bibliographystyle{plainnat}
\bibliography{references}


\clearpage
\appendix

{\LARGE
\begin{center}
\textbf{Supplementary material for ``Approximate online hierarchical clustering''}
\end{center}
}

\section{Proofs of results in main body}
\label{Proofs}
\input{proofs}

\clearpage
\section{Examples of admissible linkages [Assumption~\ref{assumption:admissible}]}
\label{Admissible}
\input{admissible}

\clearpage
\section{OTD algorithm: formal statement}
\label{sec:otd_details}
\input{otd_details}

\clearpage
\section{OHAC algorithm: formal statement}
\label{sec:ohac_details}
\input{ohac_details}

\clearpage
\section{Additional experiments}
\label{sec:additional-experiments}
\input{additional_experiments}
\end{document}

%% file: abstract.tex

Hierarchical clustering is a 
widely used approach for clustering datasets at multiple levels of granularity.
Despite its popularity, 
existing algorithms such as hierarchical agglomerative clustering (HAC) are limited to the offline setting,
and thus
require the entire dataset to be available.
This prohibits their use on large datasets commonly encountered in modern learning applications.
In this paper, we consider hierarchical clustering in the \emph{online} setting, where points arrive one at a time.
We propose two algorithms that seek to optimize the \emph{Moseley and Wang} (\emph{MW}) revenue function, 
a variant of the \emph{Dasgupta cost}.
These algorithms offer different tradeoffs between efficiency and MW revenue performance.
The first algorithm, \emph{OTD},
is a highly efficient
Online Top Down algorithm 
which provably achieves a $\frac{1}{3}$-approximation to the MW revenue under a data separation assumption.
The second algorithm, \emph{OHAC},
is
an online counterpart to offline HAC,
which
is known to
yield a $\frac{1}{3}$-approximation to the MW revenue,
and produce good quality clusters in practice.
We show that OHAC approximates offline HAC by leveraging a novel split-merge procedure.
We empirically show that 
OTD and OHAC 
offer significant
efficiency and cluster quality gains respectively
over
baselines.

%% file: intro.tex


Clustering is a canonical unsupervised learning problem, concerned with the goal of identifying groups (or \emph{clusters}) of similar instances in a dataset~\citep{Jain:1999,Aggarwal:2014}.
Real-world applications of clustering bring several challenges:
datasets may have 
an unknown number of clusters
with arbitrary (e.g., non-convex) shapes
which drift over time.
Further, individual data points may only be available in a streaming manner, rather than all at once (e.g., network traffic records),
which is known as the \emph{online} setting.
In addressing these challenges, it is further desirable to have a rigorous objective function that one's algorithm (approximately) optimizes.

Ideally, then, one would like clustering algorithms that are:
\begin{itemize}[itemsep=-2pt,topsep=-2pt]
    \item[---] \emph{non-parametric}, with no distributional assumptions about the shape or number of clusters;
    
    
    \item[---] efficiently trainable in \emph{online} settings, when data is available one at a time;
    and,
    
    \item[---] equipped with \emph{guarantees} of cluster quality when updated in online settings.
\end{itemize}
Existing methods fail to meet at least one of these challenges.
Many online algorithms have been designed only for parametric models,
such as $k$-means
~\citep{Silva:2013},~\citep{Aggarwal:2014},~\citep{Ackermann:2012b,Liberty:2014}.
Conversely, many nonparametric methods do not have a tractable online counterpart.
In particular, hierarchical agglomerative clustering (HAC)~\citep{Sneath:1973}  is a popular nonparametric
algorithm that outputs a \emph{hierarchy} or nested \emph{sequence} of clusters;
it is, however, not designed for the online setting.
While there exist online
hierarchical clustering algorithms with theoretical results~\citep{Zhang:1996,Kobren:2017},
these methods' guarantees are \emph{not} with respect to an optimal hierarchical structure.


\gui{change OTD to some other name?}
\gui{Move OTD and OHAC descriptions into bullet points?}
In this paper, we present two online hierarchical clustering algorithms that address the above challenges.
For both algorithms, we use the~\citet{Moseley:2017} (MW) revenue function (Definition~\ref{definition:moseley-rev}) to define the optimal hierarchy.
These algorithms offer different tradeoffs between efficiency and MW revenue performance.
    The first algorithm, \emph{OTD}, is an \emph{online top down} hierarchical clustering algorithm.
    OTD is very efficient as it performs a single root-to-leaf tree search when inserting a new point.
    Further, we prove that it yields a $\frac{\beta}{3}$-approximation to the MW revenue function,
    where $\beta$ parameterizes the separation of the input points.
    
    The second algorithm, 
\emph{OHAC}, 
is an online counterpart to offline HAC. 
OHAC is motivated by the fact that offline
HAC with average linkage 
in theory
yields a $\frac{1}{3}$-approximation to the MW revenue~\citep{Moseley:2017},
and in practice
produces good quality clusters.
We empirically show that OHAC,
while having more involved updates than OTD,
outperforms OTD and other baselines in terms of MW revenue.

In detail, our key contributions are as follows:
\begin{enumerate}[label=(\arabic*),itemsep=-2pt,topsep=-2pt,leftmargin=24pt]
    \item We prove that OTD yields a $\frac{\beta}{3}$-approximation to the MW revenue function (Definition \ref{definition:moseley-rev}), where $\beta$ defines the separation of the input points (Theorem~\ref{theorem:approx}).
    We provide a well-separated setting in which OTD yields a $\frac{1}{3}$-approximation, matching the \cite{Moseley:2017} offline result. 
    We experimentally show that OTD beats the $1/3$ factor on several real-world and synthetic datasets,
    while being highly efficient.
    
    \item  We experimentally show that OHAC closely approximates offline HAC in triplet distance (Definition~\ref{definition:triplet_distance})
    based on a novel splitting criterion (Definition~\ref{definition:splitting}).
    We also show that OHAC tends to outperform OTD and several baselines relative to the MW revenue function.
    
    \item For certain linkages, we guarantee a per-round time complexity of $O(d)$ for OTD, where $d$ is the depth of the hierarchy.
    Similarly, we show that OHAC has a per-round time complexity of $O(n)$ (Theorem~\ref{theorem:complexity}) for certain linkages,
    where $n$ is the number of data points.
\end{enumerate}

We proceed as follows.
In \S\ref{Background}, we review literature on offline and online hierarchical clustering,
including the Dasgupta and MW optimization view of a hierarchy.
In \S\ref{OTD}, we introduce OTD and prove the approximation factor it yields for MW revenue.
In \S\ref{ohac}, we introduce OHAC and formalize what it means to approximate offline HAC.
Lastly, in \S\ref{Evaluations}, we present experiments for OTD and OHAC.

%% file: background.tex
Our goal in this paper is to design 
online hierarchical clustering algorithms
that approximate the Moseley-Wang revenue.
Consequently, we 
first review hierarchical clustering in general;
provide the \cite{Dasgupta:2016} and \cite{Moseley:2017} optimization view of hierarchical clustering, which will serve as our central notion of cluster quality;
and review existing online hierarchical clustering approaches,
which do not aim to optimize either of these measures of cluster quality.






%
\subsection{Hierarchical clustering: definition}

\emph{Hierarchical clustering} approaches partition a given dataset into a nested \emph{sequence} of partitions,
where the number of such partitions is determined automatically~\citep{Sneath:1973,Murtagh:2012,Dasgupta:2016}. 
Within this framework, a distinction is made between top-down or divisive clustering~\citep{Kaufman:1990}, and bottom-up or hierarchical agglomerative clustering (HAC).
Hierarchical clustering approaches 
are particularly appealing
in scenarios such as exploratory data analysis, where it is unrealistic to \emph{a priori} specify the number of clusters.

In order to describe hierarchical clustering,
we first define what we mean by a hierarchy and a cluster.

\begin{definition}
\label{definition:hierarchy}
Given a set of points $X = \{ x_i \}_{i = 1}^n$,
a \emph{hierarchy} $T(X)$ over $X$ is a binary tree with $n$ leaf nodes, one for each $x_i \in X$.
A \emph{cluster} is any subset $C \subseteq X$.
\end{definition}


Given a hierarchy $T(X)$, each intermediate node $z$ induces a cluster
whose elements are all leaf nodes falling in the subtree with root $z$.
Consequently,
any hierarchy $T(X)$ implicitly defines a sequence of clusters over $X$; thus we use the terms ``hierarchy'' and ``hierarchical clustering'' interchangeably.




\subsection{Hierarchical clustering: objectives}

In \citet{Dasgupta:2016}, a cost function was introduced which defines an optimal hierarchy: given a tree $T(X)$ on $n$ input points,
and a matrix $W(X) = [ w_{ij} ]_{i, j = 1}^n$ of similarities between points: 
\begin{definition}[Dasgupta cost]
\label{definition:dasgupta-cost}
The Dasgupta cost for a hierarchy $T$ and weights $W$ is
$$ \mathrm{Cost}(T;W) = \sum_{1 \leq i \leq j \leq n} w_{ij} \cdot |\leaves(\lca(i,j))|, $$
where $w_{ij}$ is the similarity between points $i$ and $j$, $\leaves(v)$ is the set of leaves in the tree rooted at $v$, and $\lca(i, j)$ is the least common
ancestor of leaves $i$ and $j$ in $T$.    
\end{definition}

The intuition behind this cost function is the following: for a pair of leaves $i$ and $j$ that are highly similar, a ``good'' hierarchy $T$ would place them close together in the tree, which would be reflected in $\leaves(\lca(i, j))$ having small cardinality. 
In \cite{Moseley:2017}, a complementary notion of Moseley and Wang \emph{revenue} is introduced.
\begin{definition}[Moseley and Wang revenue]
\label{definition:moseley-rev}
The Moseley and Wang revenue for a hierarchy $T$ is
$$ \Rev(T;W) = \sum_{1\leq i < j\leq\ n} w_{ij} \cdot |\leaves(\lca(i, j))^{c}|, $$
\end{definition}

where $S^c = X \setminus S$ for a set of input points $S \subseteq X$. Note that for any hierarchy $T(X)$ on $n$ points, $\Cost(T;W) + \Rev(T;W) = n\sum_{1\leq i < j\leq n}w_{ij}$. Thus, the hierarchy minimizing the Dasgupta cost also maximizes the Moseley and Wang revenue.

\citet{Moseley:2017} show that offline hierarchical agglomerative clustering (HAC) with average linkage achieves a $\frac{1}{3}$-approximation to Definition~\ref{definition:moseley-rev}.\footnote{Interestingly, \cite{Charikar:2019} show that a random tree achieves the same approximation factor and give a semidefinite programming based algorithm that achieves a 0.336379-approximation.}
We now delve deeper into offline HAC.

\subsection{Offline hierarchical clustering algorithms: HAC}

Hierarchical agglomerative clustering (HAC) is a popular hierarchical clustering algorithm. 
Informally, HAC begins by treating each data point as a separate cluster.
One then iteratively selects the two \emph{maximally similar} clusters to merge into a new cluster, until all points belong in one cluster.
Appealingly, HAC with average linkage yields a $\frac{1}{3}$-approximation to
the Moseley-Wang revenue. 







Despite its merits, HAC does not scale well:
for a general linkage, the fastest implementations run in time $O(n^2 \log n )$~\citep{Day:1984} on $n$ data points.
Furthermore, HAC is not suited to 
the online setting,
where data points are available one at a time.
These weaknesses motivate the study of scalable, \emph{online} versions of HAC that can also approximate the Moseley-Wang revenue.





%
\subsection{The online hierarchical clustering problem}

We now
define the online hierarchical clustering problem of interest in this paper.
In the online clustering setting,
there is an infinite stream of points $\{ x_i \}_{i=1}^{\infty}$, with each $x_i\in\mathbb{R}^d$.
In consecutive rounds $k = 0, 1, 2, \ldots$, we observe a new point $x_{k+1}$, and must maintain a hierarchy $T_k$ over the points $X_k := \{ x_1, \ldots, x_k \}$ seen so far. Thus, an online hierarchical clustering algorithm must produce a new hierarchy $T_{k+1}$ that incorporates this point.

From an optimization lens, in the case that we have ground truth similarity measures $W_k = (w_{ij})_{1\leq i, j\leq k}$
with $w_{ij}$ denoting the similarity between $x_i$ and $x_j$, we would like
the resulting Moseley-Wang revenue $\mathrm{Rev}( T_{k + 1}; W_{k+1} )$ 
to be (approximately) maximal \footnote{When ground truth distances are provided, we may transform these to similarities by e.g. negating them.}.

Online hierarchical clustering has received limited attention, but with some exceptions.
These include algorithms that process batch data in an online manner for efficiency~\citep{Sun:2009,Loewenstein:2008,Nguyen:2014}, and an online top-down algorithm
\gui{sounds like OTD. should change wording or rename OTD}~\citep{Rodrigues:2006}.
Two particularly relevant methods are BIRCH \citep{Zhang:1996} and PERCH \citep{Kobren:2017}.

BIRCH~\citep{Zhang:1996} is a top-down rather than bottom-up clustering approach.
A BIRCH tree can have branching factor greater than two (user specified), and its leaves are themselves clusters (whose maximum size and diameter are also user specified).
PERCH~\citep{Kobren:2017} was recently proposed for ``extreme clustering'' problems, characterized by large numbers of samples and clusters.
Here, it is assumed that there is a ground truth flat clustering of the data,
and the goal is to infer a hierarchical clustering which respects \emph{dendrogram purity}: a distance function between a hierarchical clustering and a flat clustering of the same dataset.

Both our proposed algorithms are different from these works in  key respects:
\begin{enumerate}[label=(\arabic*),itemsep=0pt,topsep=0pt,leftmargin=24pt]
    \item Our OTD algorithm provides a
guarantee 
with respect to
an optimal \emph{hierarchy}, per Definition~\ref{definition:moseley-rev}.
By contrast,
PERCH provides a guarantee 
with respect to
an optimal \emph{flat} clustering.

    \item Our OHAC algorithm is algorithmically different from both methods.
Compared to BIRCH,
our tree splits can change the tree drastically compared to BIRCH's  incremental changes.
Compared to PERCH, the objective function is fundamentally different.
The aim of OHAC is to approximate the output of offline HAC at each step, which would approximate an optimal hierarchy, per Definition~\ref{definition:moseley-rev}.
By contrast, PERCH is agnostic to the precise hierarchy generated, as long as it agrees with the underlying ground truth flat clustering.
\end{enumerate}

%% file: otd.tex

We present OTD, our first algorithm for online hierarchical clustering.
This algorithm performs highly efficient online updates,
and provably approximates the Moseley-Wang revenue.


\subsection{The OTD algorithm}

Our OTD algorithm
performs an online top down update.
To proceed, we need the following
notions of average inter- and intra-subtree similarity
for subtrees $A, B$ of a hierarchy $T$:
\begin{align*}
\wbar(A) = \frac{1}{\binom{|A|}{2}} \cdot \sum_{\substack{i, j\in \leaves(A) \\ i < j}} w_{ij}
\quad
\text{ and }
\quad
\wbar(A, B) = \frac{1}{|A| \cdot |B|} \sum_{i \in \leaves(A)} \sum_{j \in \leaves(B)} w_{ij}.
\end{align*}


Given a hierarchy $T_k$ created from the first $k$ points $X_k$ and a new point $x_{k+1}$, OTD updates $T_k$ by first comparing the average similarity of pairs of points in $T_k$, $\wbar(T_k)$ with the average similarity of $x_{k+1}$ to $T_k$, i.e.
$\wbar(T_k, \{x_{k+1}\})$.
We proceed depending on the outcome of this comparison:
\begin{enumerate}[label=(\roman*),itemsep=0pt,topsep=0pt,leftmargin=16pt]
    \item If the former quantity is larger, OTD outputs $T_{k+1}$ whose children are $x_{k+1}$ and $T_k$.

    \item Otherwise, denoting the children of $T_k$ by $A_k$ and $B_k$, OTD compares the average similarities of the new point to each of these subtrees, i.e., $\wbar(A_k, \{x_{k+1}\})$ with $\wbar(B_k, \{x_{k+1}\})$.
    Supposing without loss of generality that $\wbar(B_k, \{x_{k+1}\}) \geq \wbar(A_k, \{x_{k+1}\})$, OTD proceeds by recursively inserting $x_{k+1}$ into $B_k$.
\end{enumerate}
The updates in both cases make sense intuitively.
For case (i), if the new point is very dissimilar to the previously seen points, it should be placed far away from the existing point, which is achieved by attaching it as a sibling of the existing hierarchy.
Case (ii) is similarly a reasonable greedy choice for proceeding in a top down fashion.
The algorithm update is illustrated in Figure~\ref{fig:otd-illustration};
see Appendix~\ref{sec:otd_details} for a formal description of the algorithm.

\begin{figure*}[!t]
    \centering
    \label{fig:otd-illustration-main}
    \subfloat[Existing hierarchy and new point $x$.]{
        \includegraphics[scale=0.95]{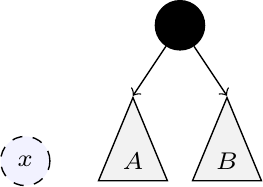}
        \label{fig:otd-illustration-existing}
    }
    \qquad
    \subfloat[Case (i).]{
        \includegraphics[scale=0.95]{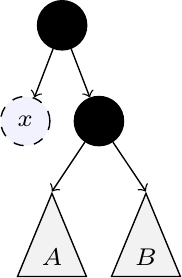}
        \label{fig:otd-illustration-case1}
    }
    \qquad
    \subfloat[Case (ii).]{
        \includegraphics[scale=0.95]{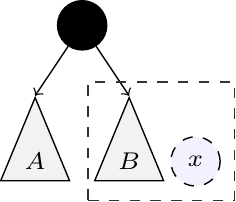}
        \label{fig:otd-illustration-case2}
    }

    \caption{Illustration of the OTD algorithm execution given an existing tree, and a new query point $x$ (panel (a)).
    One compares the average similarity of the point to the current tree against the average similarity of points in the current tree.
    If the former is larger (Case (i)), we attach $x$ as a sibling to the tree.
    If the former is smaller (Case (ii)), we find the subtree which $x$ is more similar to (in the illustration, $B$),
    and recursively perform the OTD update.
    }
    \label{fig:otd-illustration}
    \vspace{-\baselineskip}
\end{figure*}

\subsection{Approximation guarantee for OTD}

A salient aspect of OTD is that we can provide a provable approximation guarantee for the Mosley-Wang revenue.
This guarantee relies on the following \emph{well-separatedness} assumption of the data.
Many existing guarantees of clustering quality rest on similar assumptions; see, e.g.,~\citet{Kobren:2017} for an analogous notion of \emph{flat} separability.

\begin{assumption}[$\beta$-well-separated]
\label{assumption-well-separated}
Let $T_n$ be a hierarchy on $X_n$.
We say that $T_n$ is \emph{$\beta$-well-separated} ($0 \leq \beta \leq 1$) if for every subtree $S$ of $T_n$ with children $A$ and $B$, and new point $x_{n+1}$,
$$ \wbar(S, \{x_{n+1}\}) > \wbar(S) \land \wbar(A, \{x_{n+1}\}) \leq \wbar(B, \{x_{n+1}\}) \implies \wbar(A) \geq \beta\wbar(A, \{x_{n+1}\}). $$
\end{assumption}

Under this assumption, we have the following guarantee for OTD (full proof in Appendix~\ref{Proofs}).
\begin{theorem}
\label{theorem:approx}
If Assumption \ref{assumption-well-separated} is satisfied with parameter $\beta$, Algorithm \ref{algorithm:otd} is a $\frac{\beta}{3}$-approximation algorithm for the Moseley-Wang revenue (Definition~\ref{definition:moseley-rev}).
\end{theorem}

\begin{proof}[Proof sketch (full proof in Appendix)]
Let $w(A) = \wbar(A) \cdot {\binom{|A|}{2}}$ and $w(A, B) = \wbar(A, B) \cdot {|A| \cdot |B|}$.
Observe that an upper bound on the revenue of $T_n$ is $(n-2)\sum_{1\leq i\leq j\leq\ n}w_{ij}$.
One can show that the maximum revenue gain when encountering point $x_{n+1}$ is $\Delta^{\text{Max}}_{n+1} = w(T_n) + (n-1)w(T_n, \{x_{n+1}\})$.
We can then break the proof into the two cases described in the algorithm and Figure~\ref{fig:otd-illustration},
comparing the revenue gain of OTD when encountering $x_{n+1}$ ($\Delta^{\OTD}_{n+1}$), to $\Delta^{\text{Max}}_{n+1}$.

\textbf{Case i.} $\wbar(T) \geq \wbar(T, \{x_{n+1}\}) \implies w(T) \geq ((n-1)/2)w(T, \{x_{n+1}\})$ [see Figure~\ref{fig:otd-illustration-case1}].

In this case, we can show that $\Delta^{\text{Max}}_{n+1} = w(T) + (n-1)w(T, \{x_{n+1}\}) \leq w(T) + 2w(T) = 3\Delta^{\OTD}_{n+1}$.

\textbf{Case ii.} $\wbar(T, \{x_{n+1}\}) > \wbar(T) \implies ((n-1)/2)w(T, \{x_{n+1}\}) > w(T)$ [see Figure~\ref{fig:otd-illustration-case2}].

We first prove by induction,
using Assumption~\ref{assumption-well-separated},
that $\Delta^{\OTD}_{n+1} \geq (\beta(n-1)/2)w(T, \{x_{n+1}\})$.

Then, $\Delta^{\text{Max}}_{n+1} = w(T) + (n-1)w(T, \{x_{n+1}\}) < 3((n-1)/2)w(T, \{x_{n+1}\}) \leq (3/\beta)\Delta^{\OTD}_{n+1}$.
\end{proof}

\vspace{-4mm}
Assumption \ref{assumption-well-separated} is a way of imposing a hierarchical structure on the data.
For each pair of sibling subtrees $A$ and $B$, Assumption \ref{assumption-well-separated} requires that if any new $x_{n+1}$ is closer on average to $B$ than $A$, then it is significantly far from $A$, in the sense that the average pairwise similarity within points in $A$ is greater than the average similarity between $x_{n+1}$ and $A$.
Our assumption has the advantage of being parameterized, with $\beta$ quantifying the extent to which it is satisfied. As we allow $\beta$ to decrease, the assumption is satisfied by a larger class of (possibly noisy) datasets. 

\textbf{Complexity of OTD.} The fact below (explanation in Appendix~\ref{sec:otd_details}) states that OTD performs efficient updates.
\begin{fact}[OTD Complexity]
\label{fact:otd-complexity}
Consider OTD applied to dataset $X=\{x_i\}_{i=1}^n$ with linkage that can be computed in $O(1)$.
For any round $k$, $T_k$ can be computed in time $O(d)$, where $d$ is the depth of $T_k$.
\end{fact}

We provide a discussion on the variety of linkages that can be computed in $O(1)$ time in Appendix~\ref{sec:otd_details}.

\gui{proof isn't much of a ... proof. More like a set of explanations.}



%% file: ohac.tex

We now present our second online hierarchical clustering algorithm.
While OTD greedily maximized the MW revenue (Definition~\ref{definition:moseley-rev}),
here we explore another avenue:
since
offline HAC yields a $\frac{1}{3}$-approximation to the MW revenue,
we seek
an online algorithm which approximates offline HAC (and in turn, the MW revenue).
The resulting OHAC algorithm provides a different operating point to OTD:
while less efficient, it inherits offline HAC's good cluster quality,
as we show empirically.

To proceed, we first formally define the sense in which we seek to approximate offline HAC.



%
\subsection{The triplet distance between hierarchies}

Our goal is to design an online counterpart to offline HAC;
fundamentally, such an algorithm should output a hierarchy which is ``close'' to that of offline HAC.
To measure ``closeness'' of two hierarchies, we will rely on the notion of the triplet distance~\citep{Emamjomeh-Zadeh:2018}.

\begin{definition}[Triplet distance]
\label{definition:triplet_distance}
Given a hierarchy $T$ on $\{x_i\}_{i=1}^n$, let 
$\mathrm{Trip}(T) := \{(i, j, k) \colon i, j, k \text{ distinct } \land x_k \notin \mathrm{leaves}(\mathrm{lca}(x_i, x_j))\}$.
The \emph{triplet distance} between $T_1, T_2$ is
\begin{equation}
	\label{eqn:triplet-distance}
	\Delta_{\mathrm{trip}}(T_1, T_2) := {|\mathrm{Trip}(T_2)\setminus \mathrm{Trip}(T_1)|}/{|\mathrm{Trip}(T_2)|}.
\end{equation}
\end{definition}


Intuitively, 
given two points $x_i, x_j$, the least common ancestor $\mathrm{lca}( x_i, x_j )$ specifies the most fine-grained cluster which contains them.
Now, given three points, exactly one of them does not belong to the cluster induced by the LCA of the other two.
This information is what $\mathrm{Trip}(T)$ captures, and in fact two hierarchies are equivalent if and only if they have the same triplets.



Given the above, we 
can precisely state our goal of approximating HAC.
Suppose we have seen $k$ points in the online stream.
Let $T^*_k$ be the solution of offline HAC applied to these $k$ points,
and let $T_k$ be the hierarchy produced by some online algorithm.
Our goal is to design an algorithm such that $\Delta_{\mathrm{trip}}( T_k, T^*_k )$ is small for each $k = 1, 2, \ldots$.

    
    
        


%
\subsection{Online approximation of offline HAC: a general recipe}

We now present a general recipe to approximate the offline HAC tree (in a triplet distance sense) in an online setting.
Given the current hierarchy $T_k$ and new point $x_{k+1}$, one proceeds to:
\begin{enumerate}[itemsep=0pt,topsep=0pt,label={(\arabic*)}]
    \item break the hierarchy $T_k$ based on a \emph{splitting procedure} into set of subtrees (or \emph{forest}); 
    \item add the singleton tree with node $x_{k+1}$ to the forest;
    \item apply $\mathrm{HAC}$ to the new forest and merge the trees to produce the updated hierarchy $T_{k+1}$.
\end{enumerate}
Intuitively, for suitable implementation of step (1), this should result in a tree which minimizes the triplet distance to the offline HAC tree. One can imagine leaving subtrees that are far away from the new point intact, while splitting those that have leaves close to the new point.

We thus arrive at the question of determining a suitable splitting procedure.
To do so, it is instructive to consider two na\"{i}ve algorithms that operate on extreme points of the accuracy-efficiency curve.
In the first algorithm, \textbf{Na\"{i}ve$_1$}, we re-run offline HAC on the entire dataset for a new point $x_{k+1}$.
This perfectly optimizes $\Delta_{\mathrm{trip}}( T_k, T^*_k )$,
but is inefficient, with per-round time complexity $O(k^2\log k)$.

\begin{figure*}[t]
    \centering


    \subfloat[Existing hierarchy and new point $x$.]{\includegraphics[scale=0.85]{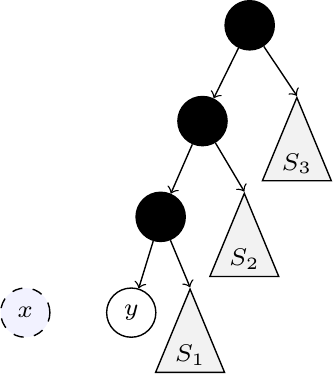}}
    \qquad
    \subfloat[Breaking hierarchy into a forest.]{\includegraphics[scale=0.85]{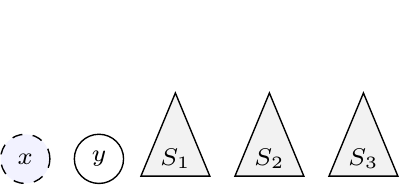}}
    \qquad
    \subfloat[Result of Na\"{i}ve$_2$.]{\includegraphics[scale=0.85]{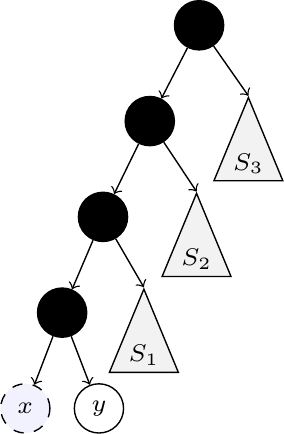}}
    \qquad
    \subfloat[Result of OHAC.]{\includegraphics[scale=0.85]{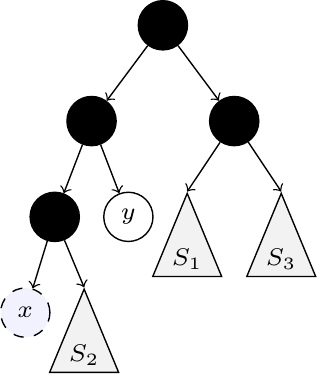}}

    \caption{Illustration of online algorithms' execution given an existing tree, and a new query point $x$ (panel (a)).
    The first step is to 
    find the nearest neighbour leaf node $y$, and
    break the hierarchy into a forest of subtrees (panel (b)).
    In {\bf Na\"{i}ve$_2$} (panel (c)), we create a new subtree with $x$ and $y$ as leaves.
    In {\bf OHAC} (panel (d)), we re-do hierarchical clustering on all subtrees ending at the root node.
    Panel (d) illustrates 
    that the resulting tree may be markedly different in structure than the original.}
    \label{fig:algorithm}
    \vspace{-\baselineskip}
\end{figure*}

In the second algorithm, \textbf{Na\"{i}ve$_2$}, we find the nearest neighbor $y_{k+1}$ of $x_{k+1}$ from the set $X_k = \{ x_i \}_{i = 1}^k$. Intuitively, as this point is the most similar to $x_{k + 1}$, the two should belong to the same cluster. The next step, then, is to simply replace the leaf node corresponding to $y_{k+1}$ with a subtree containing both $x_{k+1}$ and $y$ as nodes.
This process is visualized in Figure~\ref{fig:algorithm} (a) and (c). 
This process is efficient, but can lead to a poor worst-case approximation to $T^*_k$.
Consider three points on the line,
at positions $0$, $1$, and $K$.
If they arrive sequentially, this algorithm merges point $K$ with point $1$ first in the hierarchy, leading to arbitrarily poor MW revenue approximation for large $K$. 


%


In terms of the above recipe, step (1) of Na\"{i}ve$_1$ chooses to break $T_k$ maximally into singleton trees, one for each leaf node.
On the other hand, step (1) of Na\"{i}ve$_2$ minimally breaks $T_k$, at the cost of a poor worst-case approximation of $T^*_k$.
We now present our algorithm,
OHAC,
which aims to improve on the efficiency of {Na\"{i}ve$_1$}, while still being as faithful as possible to the output of offline HAC.

\subsection{Online approximation of offline HAC: the OHAC algorithm}

The OHAC algorithm employs the following splitting procedure to choose a suitable set of subtrees to split the hierarchy $T_k$ in step 1. 
See also Figure~\ref{fig:algorithm}(a) and (b).

\begin{definition}[Splitting]
\label{definition:splitting} 
Given a hierarchy $T$ and leaf node $y$, define $\mathrm{Split}(T, y)$ as the set of subtrees
\begin{center}
$S_0 = \{y\}$ \hspace{5mm}
$S_1 = \mathrm{Sibling}(y)$ \hspace{5mm}
$S_i = \mathrm{Sibling}(\mathrm{Parent}(S_{i-1})), i = 2, \ldots, \mathrm{depth}({y})$
\end{center}
\vspace{-2mm}
\end{definition}

Note that $\bigcup_{S\in \mathrm{Split}(T, y)} \mathrm{leaves}({S}) = X$, so the $S_i$'s partition $X$.
The OHAC algorithm (Algorithm \ref{algorithm:OHAC}) applies the above procedure to split the nearest neighbor $y_{k+1}$ of the newly inserted point $x_{k+1}$.
While Na\"{i}ve$_2$ merges $x_{k+1}$ with $y_{k+1}$, it is possible that $y_{k+1}$'s nearest neighbor is not $x_{k+1}$ but its sibling $S_1$.
Similarly, $S_1$'s nearest neighbor might be its sibling $S_2$, and not $x_{k+1}$.
Thus, we consider the \emph{whole} set of subtrees $S_1, \ldots, S_{\mathrm{depth}(y_{k+1})}$ as possible merge candidates for $x_{k+1}$.

Finally, it is possible that once $x_{k+1}$ is merged with the appropriate subtree, the distances between the newly created subtree and other subtrees are changed "higher up" in the hierarchy.
We address this potentiality by applying $\mathrm{HAC}$ to the forest.
Unlike Na\"{i}ve$_1$, which applies HAC to the trivial forest of singleton subtrees, note that we apply HAC to a relatively small number of inputs. This is quantified in the next section where we show how our splitting procedure is computationally efficient.
\subsection{Complexity analysis of OHAC}
\label{complexity}

We derive the computational complexity of OHAC based on two assumptions.
The first is a technical condition on the hierarchies we encounter in the online process.

\begin{assumption}[Balance]
\label{assumption:balancedness}
A hierarchy $T$ with $k$ leaves is \emph{balanced} if $\mathrm{height}(T) = O(k^{\frac{1}{2}}/ \log k)$.
\end{assumption}


\begin{assumption}[Admissible linkage]
\label{assumption:admissible}
Call a linkage $L$ \emph{admissible} if $\exists$ an algorithm which given clusters $C_1, \ldots, C_m$ computes $\{L(C_i, C_j)\}_{i\neq j}$ in time $O(m^2+\sum_{i=1}^m|C_i|)$.
\end{assumption}

Centroid linkage, average linkage with dot product, and moment-based linkages are all
admissible, showing that this class is not unduly restrictive.
For details of these linkages, see Appendix~\ref{Admissible}.

Under Assumptions \ref{assumption:balancedness} and \ref{assumption:admissible}, we have:
\begin{theorem}[OHAC Complexity]
\label{theorem:complexity}
Consider OHAC applied to dataset $X=\{x_i\}_{i=1}^n $ with respect to an admissible linkage $L$. Then, for any round $k$, if $T_{k-1}$ is balanced, $T_k$ is computed in time $O(k)$.
\end{theorem}

%% file: evaluation.tex

We now present empirical results 
verifying that OTD and OHAC achieve their core aim of
providing efficient online algorithms
that reliably approximate the MW revenue.
Further, we verify that they provide different tradeoffs:
while OTD is significantly more efficient,
OHAC is more accurate,
including against the baselines of offline HAC and PERCH~\citep{Kobren:2017}.



%
We evaluate the algorithms on both synthetic and real-world datasets.
For the former,
we construct {\tt gmm-1}, {\tt gmm-8}, and {\tt gmm-100}, three synthetic datasets comprising points drawn from a mixture of $1$, $8$, and $100$ clusters respectively.
For the latter, we use the UCI datasets {\tt iris}, {\tt glass}, and {\tt mnist} as well as ILSVRC12 projected down by PCA to allow for runtimes in hours.
See Table~\ref{tbl:dataset} in Appendix~\ref{sec:additional-experiments} for detailed statistics of the datasets.
For each dataset, 
we run offline HAC on the entire data, and each online algorithm by processing the data points one at a time. 

\begin{figure}[!t]
  \centering
  \includegraphics[scale=0.2]{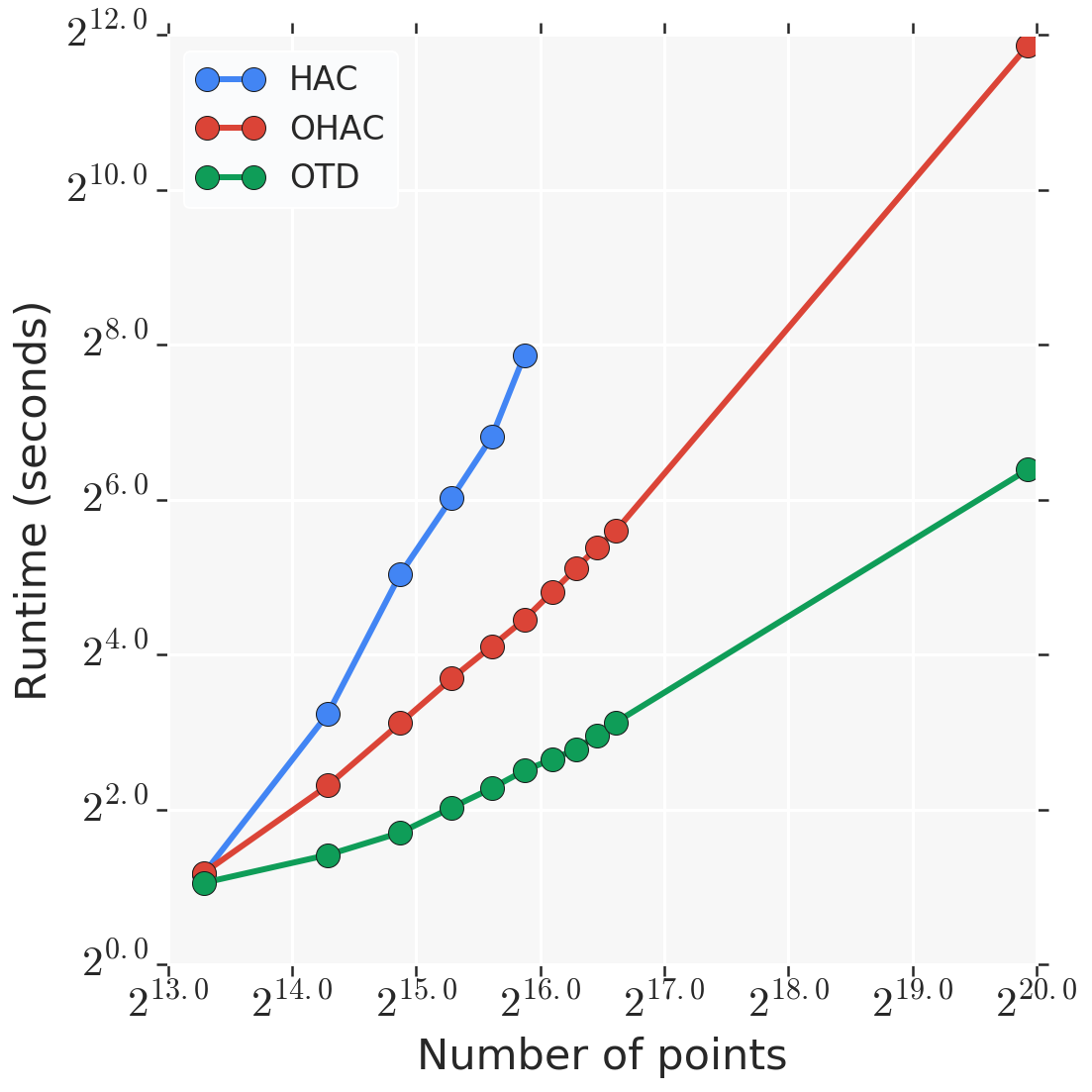}
  \caption{Runtime of OHAC, OTD, and offline HAC versus dataset size.
  Offline HAC runs out of memory at 70K points.
           The dataset consists of uniformly generated points within the unit square.}
  \label{figure:scale}
  \vspace{-\baselineskip}
\end{figure}

%
Table \ref{tbl:MW-revenue} compares the MW Revenue for all methods.
\footnote{Performance numbers are approximated by sampling, since exhaustive computation is prohibitive.} 
We draw the following conclusions:
\begin{enumerate}[label=(\arabic*),itemsep=0pt,topsep=0pt,leftmargin=16pt]
    \item OTD consistently achieves, or improves upon, the approximation factor of $1/3$ that is guaranteed by our theory.
    This is despite the fact that these datasets are not $1$-well-separated;
    thus, OTD can produce good quality clusters even in regimes not covered by the theory.
    
    \item offline HAC yields the best quality clustering across the datasets.
    However, its runtime is also the most prohibitive:
    see Figure~\ref{figure:scale}, showing that it is consistently slower than OHAC and OTD.
    
    \item the approximation factors for OHAC are very similar to that of HAC, which validates our approach of approximating HAC as a way to optimize the MW Revenue.

    \item the {\tt imagenet} dataset consisting of over one million points proves overwhelming for offline HAC. However, both OTD and OHAC are capable of finishing in a reasonable amount of time.
\end{enumerate}

\aditya{probabilistic triplet distance}

\aditya{We need a table or plot of runtime numbers.}

\begin{table}[!t]
    \centering

	\resizebox{0.99\linewidth}{!}{
	\begin{tabular}{llllllll}
		\toprule
		\texttt{} & \texttt{gmm-$1$} & \texttt{gmm-$8$} & \texttt{iris} & \texttt{glass} & \texttt{mnist} & \texttt{gmm-$100$} & \texttt{ILSVRC12}\\
		\midrule
		\midrule
HAC    & 120 (0.58) & 9.83 (0.95)   & 22.1 (0.74) & 25.4 (0.68) & 6.16K (0.36)& DNF & DNF\\
\midrule
PERCH  & 99.0 (0.48) & 9.57 (0.93)   & 19.82 (0.67) & 16.72 (0.44) & \best{6.09K (0.36)} & 173K (0.41) & 140K (0.37) \\
OTD   & 69.9 (0.34)  & 4.12 (0.40)   & 13.9 (0.47) & 20.2 (0.53) & 5.73K (0.34) & 137K (0.33) & 125K (0.33)\\
OHAC    & \best{118 (0.57)} & \best{9.81 (0.95)}  & \best{22.1 (0.74)} & \best{25.4 (0.68)} & 5.78K (0.34) & \best{182K (0.44)} & \best{153K (0.40)}\\
		\bottomrule
	\end{tabular}
	}
	
    \caption{MW Revenue normalized by number of pairs,
    with fraction of maximum revenue in parentheses; $\uparrow$ is better.
    We have highlighted in each column the best performing online method (lower panel).
    ``DNF'' indicates that a method did not finish training within 8 hours.
    }
    \label{tbl:MW-revenue}
    \vspace{-1\baselineskip}
\end{table}

We additionally compare the performance of OHAC versus offline HAC.
Our aim is to confirm that OHAC indeed approximates the HAC tree,
while being more efficient to train in an online setting.

Table \ref{tbl:triplet_distance} 
compares the hierarchies generated by OHAC and PERCH against HAC with respect to triplet distance.
For any dataset $X$, the expected distance between two random hierarchies $T_1$ and $T_2$ on $X$ is $\frac{2}{3}$, which is a baseline for interpreting the results in Table \ref{tbl:triplet_distance}. 
{\tt gmm-$8$} was designed to have well separated clusters, and both OHAC and PERCH accurately recreate the hierarchy.
On both {\tt iris} and {\tt glass}, OHAC performs relatively well in capturing the hierarchy of HAC, compared to PERCH. Finally, {\tt gmm-$1$} is a single Gaussian cluster with no sub-clusters. Even here, OHAC is able to recreate a hierarchy with triplet distance about half that of a random hierarchy. 

Finally, Figure \ref{figure:scale} plots the running times of OHAC, OTD and offline HAC as a function of dataset size. 
The runtime for the former two is the total time to process all points sequentially.
We see that OHAC is asymptotically faster than HAC,
with OTD being faster still, 
validating our claim of efficiency.

\begin{table}[!t]
    \centering
	\resizebox{0.5\linewidth}{!}{
	\begin{tabular}{llllll}
		\toprule
		\texttt{} & \texttt{gmm-$8$} & \texttt{gmm-$1$} & \texttt{iris} & \texttt{glass} & \texttt{mnist}\\
		\midrule
		\midrule
OHAC   & 0.01  & \best{0.36} & \best{0.02} & \best{0.12} & 0.44\\
PERCH  & 0.01  & 0.59 & 0.35 & 0.52 & \best{0.41} \\
		\bottomrule
		\vspace{-\baselineskip}
	\end{tabular}
	}
	%
	\caption{Triplet distance with respect to offline HAC hierarchy; $\downarrow$ is better.}
	\label{tbl:triplet_distance}
	\vspace{-\baselineskip}
\end{table}

\gui{mention which dataset is used for figure~\ref{figure:scale}}

%% file: conclusion.tex

The prevalence of streaming datasets, low latency and large throughput requirements necessitate the need for online hierarchical clustering algorithms.
In this paper, we have studied two such algorithms, OTD and OHAC, which offer operational tradeoffs between efficiency and cluster quality.
We have justified these algorithms both theoretically and experimentally, by comparing their output
to a widely accepted definition of an optimal hierarchy.
These are the first online hierarchical clustering algorithms shown to approximate an optimal hierarchy.


%% file: proofs.tex
\begin{proof}[Proof of Theorem \ref{theorem:approx}]

$ $\\ 

First, we introduce some more notation. For trees $A$ and $B$, let
\begin{enumerate}[label=(\roman*),itemsep=0pt,topsep=0pt,leftmargin=16pt]
    \item $w(A) = \sum_{\substack{i, j\in \leaves(A) \\ i < j}} w_{ij}$ (i.e. $w(A) = \wbar(A) \cdot {\binom{|A|}{2}}$).
    \item $w(A, B) = \sum_{i \in \leaves(A)} \sum_{j \in \leaves(B)} w_{ij}$ (i.e. $w(A, B) = \wbar(A, B) \cdot {|A| \cdot |B|}$)
\end{enumerate}

\gui{perhaps change $\Rev(T)$ to $\Rev(T, W)$?}


Now, note that an upper bound on the revenue gain for point $n$ is
\begin{equation}
\MaxRev_n :=(n-2)\sum_{1\leq i\leq j\leq\ n}w_{ij}.
\end{equation}

Let $T_n := \OTD(X_n)$ and $\Delta^{\OTD}_{n+1} := \Rev(T_{n+1}) - \Rev(T_n)$.

Also define 
\begin{align*}
\Delta^{\text{Max}}_{n+1} &=   \MaxRev_{n+1} - \MaxRev_n\\
&= \sum_{1\leq i\leq j\leq\ n}w_{ij} +(n-1)\sum_{i=1}^{n}w_{i, n+1}\\
&= w(T_n) + (n-1)w(T_n, \{x_{n+1}\})\\
\end{align*}
By the property of telescoping series, it suffices to show that $\Delta^{\OTD}_{n+1} \geq (\beta/3)\Delta^{\text{Max}}_{n+1}$ where $0 \leq \beta \leq 1$.




Set $T = T_n$, and let $A$, $B$ be subtrees of $T$ such that, WLOG, $\wbar(A, \{x_{n+1}\}) \leq \wbar(B, \{x_{n+1}\})$.\\
\paragraph{Case 1} $\wbar(T) \geq \wbar(T, \{x_{n+1}\})$, i.e. $w(T) \geq ((n-1)/2)w(T, \{x_{n+1}\})$.\\
\begin{equation}
\Delta^{\OTD}_{n+1} = \Rev(\Merge(T, x_{n+1})) - \Rev(T) = w(T)
\end{equation}
Now, we have
\begin{equation}
\Delta^{\text{Max}}_{n+1} = w(T) + (n-1)w(T, \{x_{n+1}\}) \leq w(T) + 2w(T) = 3\Delta^{\OTD}_{n+1}
\end{equation}

\paragraph{Case 2} $\wbar(T, \{x_{n+1}\}) > \wbar(T)$, i.e, $((n-1)/2)w(T, \{x_{n+1}\}) > w(T)$\\

In this case, we first prove by induction on the size of $T$ that
\begin{equation}
\label{equation:induction-general}
\Delta^{\OTD}_{n+1} = \Rev(\Merge(T, x_{n+1})) - \Rev(T) \geq \beta\frac{n-1}{2}w(T, \{x_{n+1}\}).
\end{equation}

\textbf{Base case.} Given $X_2 = \{a, b\}$, we show the above holds for new point $c$.

We assume WLOG that $w_{bc} \geq w_{ac}$ and thus insert $c$ as a sibling of $b$.
Now, for the LHS of inequality~\ref{equation:induction-general} we have $\Rev(\Merge(T, \{c\})) - \Rev(T) = w_{bc}$.
For the RHS of inequality~\ref{equation:induction-general}, we have $(\beta/2)w(T, \{c\}) = (\beta/2)(w_{ac} + w_{bc})$.
We know $w_{bc} \geq w_{ac}$, and thus have shown the base case to be true.

\textbf{Inductive step.} By Assumption \ref{assumption-well-separated}, 

\begin{equation}
w(A) \geq \beta\frac{|A|-1}{2}w(A, \{x_{n+1}\}) \label{equation:A-assumption}.
\end{equation}

Also, since $\wbar(B, \{x_{n+1}\})\geq \wbar(A, \{x_{n+1}\})$, we have

\begin{equation}
|A|w(B, \{x_{n+1}\}) \geq (\beta/2)(|A|w(B, \{x_{n+1}\}) + |B|w(A, \{x_{n+1}\})). \label{equation:AWB}
\end{equation}

By our inductive hypothesis, we have
\begin{equation}
\Rev(\OTD(B, x_{n+1})) - \Rev(B) \geq \beta\frac{|B|-1}{2}w(B, \{x_{n+1}\}). \label{equation:inductive-hypothesis}
\end{equation} Then we have
\begin{align*}
\Delta^{\OTD}_{n+1} &\geq w(A) + |A|w(B, \{x_{n+1}\}) + \beta\frac{|B|-1}{2}w(B, \{x_{n+1}\})\text{ (Using \ref{equation:inductive-hypothesis})}\\
&\geq \beta\frac{|A|-1}{2}w(A, \{x_{n+1}\}) + |A|w(B, \{x_{n+1}\}) + \beta\frac{|B|-1}{2}w(B, \{x_{n+1}\})\text{ (Using \ref{equation:A-assumption})}\\
&\geq \beta\frac{|A|-1}{2}w(A, \{x_{n+1}\}) + \frac{\beta}{2}(|A|w(B, \{x_{n+1}\}) + |B|w(A, \{x_{n+1}\})) \\
& \quad + \beta\frac{|B|-1}{2}w(B, \{x_{n+1}\})\text{ (Using \ref{equation:AWB})} \\
&\geq \beta\frac{n-1}{2}w(T, \{x_{n+1}\}).
\end{align*}

Finally, we have
\begin{equation}
\Delta^{\text{Max}}_{n+1} = w(T) + (n-1)w(T, \{x_{n+1}\}) < 3((n-1)/2)w(T, \{x_{n+1}\}) \leq (3/\beta)\Delta^{\OTD}_{n+1}.
\end{equation}
\end{proof}

\begin{proof}[Proof of Theorem~\ref{theorem:complexity}]
OHAC involves three key steps:
\begin{itemize}[itemsep=-2pt,topsep=0pt]
    \item finding the nearest neighbour $y_{k+1}$
    \item splitting the hierarchy with respect to the neighbour
    \item running HAC on the resulting collection of subtrees.
\end{itemize}
The first two operations have easily computable complexity: the nearest neighbor search takes $O(k)$ time,
while the splitting of $T_k$ takes $O(l)$ time, for $l = \mathrm{depth}(y_{k+1})$.

We now discuss the complexity of applying the final HAC step.
To apply HAC, we must compute the pair-wise distances between the split clusters $\{S_0, S_1, \ldots, S_l\}$.
Given that we are using an admissible linkage, this translates to a time complexity of $O(k + l^2)$.
Recalling the time complexity of HAC, we get a complexity of $O(k + l^2 + l^2\log l$) for this last step.
Noting that by our assumption $O(l^2\log l) = O(k)$ and putting the pieces together, we arrive at an $O(k)$ per-round cost of OHAC. 
\end{proof}

%% file: admissible.tex
\begin{example}[Centroid Linkage]
We note immediately that centroid linkage with any distance function is an admissible linkage. This is because one can compute the centroids of the clusters in time $O(\sum_{i=1}^m|C_i|)$ followed by the intercluster distances in time $O(m^2)$. 
\end{example}

\begin{example}[Average linkage, dot product]
\label{example:AL1}
A canonical example of an admissible linkage is average linkage with dot product which is equivalent to centroid linkage with dot product. The latter fact follows from the computation
\begin{align*}
L^{\mathrm{AL}}_{\mathrm{dot}}(C,D) &= \frac{1}{|C||D|} \sum_{x\in C, y\in D} x^{\mathrm{T}} y\\
&=\frac{1}{|C||D|}\left( \sum_{x\in C} x\right)^{\mathrm{T}} \left( \sum_{y\in D} y\right)\\
&=\mu(C)^{\mathrm{T}} \mu(D) \\
&= L^{\mathrm{CL}}_{\mathrm{dot}}(C,D).
\end{align*}
\end{example}

Centroid linkage depends on the first moments of the two clusters. The following generalizes this observation.
\begin{definition}[Moment-based linkages] 
Define the \emph{moments} of a cluster $C$ by 
\begin{align*}
    m_0(C) &:= |C| \quad \text{ and } \quad m_i(C) := \sum_{x\in C}x^{(i)},
\end{align*}
where $x^{(i)} \in \mathbb{R}^d$ is the vector with coordinates $x^{(i)}_j = (x_j)^i$, $j\in \{1, 2, \ldots, d\}$.
We say linkage $L(C, D)$ a is \emph{k-moment linkage} if there is an algorithm that computes $L(C, D)$ from $\{m_i(C), m_i(D)\}_{i=0}^k$ in time $O(k)$.  
\end{definition}

All moment-based linkages are admissible linkages: the reasoning mimics the reasoning that centroid linkages are admissible. The following is an example of a $2$-moment linkage that is equivalent to a version of average linkage.

\begin{example}[Average linkage, $l_2^2$ distance]
\label{example:AL2}
Set $d(x, y) = ||x-y||_2^2$ and $L = L^{\mathrm{AL}}_d$. It is easily verified that
\begin{equation}
L(C, D) = \frac{m_2(C)}{m_0(C)} + \frac{m_2(D)}{m_0(D)} - 2\frac{m_1(C)^{\mathrm{T}} m_1(D)}{m_0(C) \cdot m_0(D)}
\end{equation}
and hence $L$ is a $2$-moment linkage.
\end{example}
\textbf{}

%% file: otd_details.tex
We present the formal definition of the OTD algorithm in Algorithm~\ref{algorithm:otd}.

\begin{algorithm}[H]
	\caption{The OTD algorithm.}
	\label{algorithm:otd}
	\begin{algorithmic}[1]
	    \REQUIRE Current hierarchy $T$, new point $x_k$
	    \ENSURE  Hierarchy $T$
	    \IF{$k = 1$} 
	        \STATE $T \leftarrow \Node(x_1)$
	        \COMMENT{$\triangleright$ Initialize Tree}
	    \ELSIF{$k = 2$}
	        \STATE $T \leftarrow \Merge(T, \Node(x_2))$
	    \ELSIF{$|T| = 1$ \OR $\wbar(T) \geq \wbar(T, \{x_k\})$}
	        \STATE $T \leftarrow \Merge(T, \Node(x_k))$
	    \ELSE
	        \STATE $A \leftarrow T.\mathrm{left}, B \leftarrow T.\mathrm{right}$\\
	        \IF{$\wbar(A, \{x_k\}) \leq \wbar(B, \{x_k\})$}
	            \STATE $T \leftarrow \Merge(A, \OTD(B, x_{k}))$
	        \ELSE
	            \STATE $T \leftarrow \Merge(\OTD(A, x_{k}), B)$
            \ENDIF
	    \ENDIF
	\end{algorithmic}
\end{algorithm}

\textbf{OTD runtime.} Below, we explain the runtime of OTD (Algorithm~\ref{algorithm:otd}), as stated in Fact~\ref{fact:otd-complexity}.

In analyzing Algorithm \ref{algorithm:otd}, we note that in the worst case, the algorithm recurses for $O(d)$ steps, where $d$ denotes the depth of the tree.
As we move down the hierarchy to insert $x_{n+1}$, we need to compute the self-similarity of a subset with itself and with the new point.
For Euclidean data with $O(1)$ dimensions and dot product similarity, this computation can be done in $O(1)$ time by storing and updating the centroids and sizes (number of leaves) of each of the nodes. Thus, if $w_{ij} = x_i^{\mathrm{T}} x_j$, then $\wbar(A, B) = \mu(A)^{\mathrm{T}} \mu(B)$ where $\mu(T)$ denotes the centroid of the leaves of $T$.
Once $x_{n+1}$ is inserted in the tree, the centroids and sizes of all its ancestors can be updated in $O(d)$ time.

More generally, this line of reasoning can be extended to a class of similarity functions that depend only on the first $O(1)$ moments of the nodes;
see Appendix~\ref{Admissible} for a fuller discussion.

%% file: ohac_details.tex
\begin{algorithm}
	\caption{OHAC update rule}\label{algorithm:OHAC}
	\begin{algorithmic}[1]
	    \REQUIRE Current hierarchy $T_k$, new point $x_{k+1}$
	    \ENSURE  Hierarchy $T_{k+1}$
	    \IF{$k = 0$} 
	        \STATE $T_{k+1} \leftarrow \mathrm{Node}(x_0)$
	        \COMMENT{$\triangleright$ Initialize Tree}
	    \ELSE
	        \STATE $y_{k+1} \leftarrow \operatorname{argmin}_{x\in \mathrm{leaves}(T)} d(x_{k+1}, x)$
	        \STATE $S \leftarrow \mathrm{Split}(T_k, y_{k+1})$
	        \COMMENT{$\triangleright$ Definition \ref{definition:splitting}}
	        \STATE $T_{k+1} \leftarrow \mathrm{HAC}(S \cup \mathrm{Node}(x_{k+1}))$
	    \ENDIF
	\end{algorithmic}
\end{algorithm}

%% file: additional_experiments.tex
\subsection{Description of datasets}

Table~\ref{tbl:dataset} summarises statistics of the various datasets used in our experiments.

\begin{table}[!h]
    \centering
    \renewcommand{\arraystretch}{1.0}
    \begin{tabular}{llll}
        \toprule
        \textbf{Dataset} & $n$ & $d$ & $\textbf{\# clusters}$ \\
        \midrule
        \midrule
        
        {\tt gmm-1}   & $1$K & $2$ & $1$ \\
        {\tt gmm-8}   & $800$ & $3$ & $8$ \\
        {\tt iris}   & $150$ & $2$ & $3$ \\
        {\tt glass}   & $214$ & $10$ & $6$ \\
        {\tt mnist} & $55$K & $784$ & $10$ \\
        {\tt gmm-100} & $1$M & $3$ & $100$ \\
        {\tt ILSVRC12} & $1.3$M & $10$ & $1000$ \\
        \bottomrule
    \end{tabular}
    \caption{Summary of evaluation datasets in terms of \# of instances $(n)$, dimensionality $(d)$, and intrinsic \# of flat clusters.}
    \label{tbl:dataset}
\end{table}

%% file: main_arxiv.bbl
\begin{thebibliography}{19}
\providecommand{\natexlab}[1]{#1}
\providecommand{\url}[1]{\texttt{#1}}
\expandafter\ifx\csname urlstyle\endcsname\relax
  \providecommand{\doi}[1]{doi: #1}\else
  \providecommand{\doi}{doi: \begingroup \urlstyle{rm}\Url}\fi

\bibitem[Ackermann et~al.(2012)Ackermann, M\"{a}rtens, Raupach, Swierkot,
  Lammersen, and Sohler]{Ackermann:2012b}
Marcel~R. Ackermann, Marcus M\"{a}rtens, Christoph Raupach, Kamil Swierkot,
  Christiane Lammersen, and Christian Sohler.
\newblock Streamkm++: A clustering algorithm for data streams.
\newblock \emph{J. Exp. Algorithmics}, 17:\penalty0 2.4:2.1--2.4:2.30, May
  2012.
\newblock ISSN 1084-6654.

\bibitem[Aggarwal and Reddy(2014)]{Aggarwal:2014}
Charu~C. Aggarwal and Chandan~K. Reddy, editors.
\newblock \emph{Data Clustering: Algorithms and Applications}.
\newblock Chapman \& Hall\discretionary{/}{}{/}CRC Press, Boca Raton, FL, USA,
  2014.

\bibitem[Charikar et~al.(2019)Charikar, Chatziafratis, and
  Niazadeh]{Charikar:2019}
Moses Charikar, Vaggos Chatziafratis, and Rad Niazadeh.
\newblock Hierarchical clustering better than average-linkage.
\newblock In \emph{Proceedings of the Thirtieth Annual {ACM-SIAM} Symposium on
  Discrete Algorithms, {SODA} 2019, San Diego, California, USA, January 6-9,
  2019}, pages 2291--2304, 2019.

\bibitem[Dasgupta(2016)]{Dasgupta:2016}
Sanjoy Dasgupta.
\newblock A cost function for similarity-based hierarchical clustering.
\newblock In \emph{Proceedings of the Forty-eighth Annual ACM Symposium on
  Theory of Computing}, STOC '16, 2016.

\bibitem[Day and Edelsbrunner(1984)]{Day:1984}
William H.~E. Day and Herbert Edelsbrunner.
\newblock Efficient algorithms for agglomerative hierarchical clustering
  methods.
\newblock \emph{Journal of Classification}, 1\penalty0 (1):\penalty0 7--24, Dec
  1984.

\bibitem[Emamjomeh-Zadeh and Kempe(2018)]{Emamjomeh-Zadeh:2018}
Ehsan Emamjomeh-Zadeh and David Kempe.
\newblock Adaptive hierarchical clustering using ordinal queries.
\newblock In \emph{Proceedings of the Twenty-Ninth Annual ACM-SIAM Symposium on
  Discrete Algorithms}, SODA '18, 2018.

\bibitem[Jain et~al.(1999)Jain, Murty, and Flynn]{Jain:1999}
A.~K. Jain, M.~N. Murty, and P.~J. Flynn.
\newblock Data clustering: A review.
\newblock \emph{ACM Comput. Surv.}, 31\penalty0 (3):\penalty0 264--323,
  September 1999.
\newblock ISSN 0360-0300.

\bibitem[Kaufman and Rousseeuw(1990)]{Kaufman:1990}
L.~Kaufman and P.J. Rousseeuw.
\newblock \emph{Finding Groups in Data: an introduction to cluster analysis}.
\newblock Wiley, 1990.

\bibitem[Kobren et~al.(2017)Kobren, Monath, Krishnamurthy, and
  McCallum]{Kobren:2017}
Ari Kobren, Nicholas Monath, Akshay Krishnamurthy, and Andrew McCallum.
\newblock A hierarchical algorithm for extreme clustering.
\newblock In \emph{Proceedings of the 23rd ACM SIGKDD International Conference
  on Knowledge Discovery and Data Mining}, KDD '17, pages 255--264, New York,
  NY, USA, 2017. ACM.

\bibitem[Liberty et~al.(2014)Liberty, Sriharsha, and Sviridenko]{Liberty:2014}
Edo Liberty, Ram Sriharsha, and Maxim Sviridenko.
\newblock An algorithm for online k-means clustering.
\newblock \emph{CoRR}, abs/1412.5721, 2014.

\bibitem[Loewenstein et~al.(2008)Loewenstein, Portugaly, Fromer, and
  Linial]{Loewenstein:2008}
Yaniv Loewenstein, Elon Portugaly, Menachem Fromer, and Michal Linial.
\newblock Efficient algorithms for accurate hierarchical clustering of huge
  datasets: tackling the entire protein space.
\newblock In \emph{ISMB}, 2008.

\bibitem[Moseley and Wang(2017)]{Moseley:2017}
Benjamin Moseley and Joshua Wang.
\newblock Approximation bounds for hierarchical clustering: Average linkage,
  bisecting k-means, and local search.
\newblock In I.~Guyon, U.~V. Luxburg, S.~Bengio, H.~Wallach, R.~Fergus,
  S.~Vishwanathan, and R.~Garnett, editors, \emph{Advances in Neural
  Information Processing Systems 30}. 2017.

\bibitem[Murtagh and Contreras(2012)]{Murtagh:2012}
Fionn Murtagh and Pedro Contreras.
\newblock Algorithms for hierarchical clustering: an overview.
\newblock \emph{Wiley Interdisciplinary Reviews: Data Mining and Knowledge
  Discovery}, 2\penalty0 (1):\penalty0 86--97, 2012.

\bibitem[Nguyen et~al.(2014)Nguyen, Schmidt, and Kwoh]{Nguyen:2014}
Thuy-Diem Nguyen, Bertil Schmidt, and Chee-Keong Kwoh.
\newblock Sparsehc: A memory-efficient online hierarchical clustering
  algorithm.
\newblock \emph{Procedia Computer Science}, 29:\penalty0 8 -- 19, 2014.
\newblock 2014 International Conference on Computational Science.

\bibitem[Rodrigues et~al.(2006)Rodrigues, Gama, and Pedroso]{Rodrigues:2006}
Pedro~Pereira Rodrigues, Jo{\~{a}}o Gama, and Jo{\~{a}}o~Pedro Pedroso.
\newblock {ODAC:} hierarchical clustering of time series data streams.
\newblock In \emph{Proceedings of the Sixth {SIAM} International Conference on
  Data Mining, April 20-22, 2006, Bethesda, MD, {USA}}, pages 499--503, 2006.

\bibitem[Silva et~al.(2013)Silva, Faria, Barros, Hruschka, Carvalho, and
  Gama]{Silva:2013}
Jonathan~A. Silva, Elaine~R. Faria, Rodrigo~C. Barros, Eduardo~R. Hruschka,
  Andr{\'e} C. P. L. F.~de Carvalho, and Jo\~{a}o Gama.
\newblock Data stream clustering: A survey.
\newblock \emph{ACM Comput. Surv.}, 46\penalty0 (1):\penalty0 13:1--13:31, July
  2013.
\newblock ISSN 0360-0300.

\bibitem[Sneath and Sokal(1973)]{Sneath:1973}
P.H.A. Sneath and R.R. Sokal.
\newblock \emph{Numerical Taxonomy: The Principles and Practice of Numerical
  Classification}.
\newblock Freeman, San Francisco, 1973.

\bibitem[Sun et~al.(2009)Sun, Cai, Liu, Yu, Farrell, Mckendree, and
  Farmerie]{Sun:2009}
Yijun Sun, Yunpeng Cai, Li~Liu, Fahong Yu, {Michael L.} Farrell, William
  Mckendree, and William Farmerie.
\newblock Esprit: Estimating species richness using large collections of 16s
  rrna pyrosequences.
\newblock \emph{Nucleic Acids Research}, 37\penalty0 (10), 2009.
\newblock ISSN 0305-1048.

\bibitem[Zhang et~al.(1996)Zhang, Ramakrishnan, and Livny]{Zhang:1996}
Tian Zhang, Raghu Ramakrishnan, and Miron Livny.
\newblock Birch: An efficient data clustering method for very large databases.
\newblock In \emph{Proceedings of the 1996 ACM SIGMOD International Conference
  on Management of Data}, SIGMOD '96, pages 103--114, New York, NY, USA, 1996.
  ACM.
\newblock ISBN 0-89791-794-4.

\end{thebibliography}
